\documentclass[runningheads]{llncs}

\usepackage{graphicx}
\usepackage{algorithm,algorithmic}
\usepackage{multicol}
\usepackage{amsmath}
\usepackage{multirow}
\usepackage{caption}   
\usepackage{color}
\usepackage{subfigure} 
\begin{document}
\title{M$^3$FGM:A Node Masking and Multi-granularity Message passing-based Federated Graph Model for Spatial-Temporal Data Prediction}
%
%
\author{Yuxing Tian\inst{1}\and Jiachi Luo\inst{1}\and  Zheng Liu\inst{2}\and Song Li\inst{3}\and Yanwen Qu\inst{1}}

\institute{Jiangxi Normal University\and Nanjing University of Posts and Telecommunications\and Shanghai enflame technology co. ltd}
%
\maketitle              
\begin{abstract}
Researchers are solving the challenges of spatial-temporal prediction by combining Federated Learning (FL) and graph models with respect to the constrain of privacy and security. In order to make better use of the power of graph model, some researchs also combine split learning(SL). However, there are still several issues left unattended: 1) Clients might not be able to access the server during inference phase; 2) The graph of clients designed manually in the server model may not reveal the proper relationship between clients. This paper proposes a new GNN-oriented split federated learning method, named node {\bfseries M}asking and {\bfseries M}ulti-granularity {\bfseries M}essage passing-based Federated Graph Model (M$^3$FGM) for the above issues. For the first issue, the server model of M$^3$FGM employs a MaskNode layer to simulate the case of clients being offline. We also redesign the decoder of the client model using a dual-sub-decoders structure so that each client model can use its local data to predict independently when offline. As for the second issue, a new GNN layer named Multi-Granularity Message Passing (MGMP) layer enables each client node to perceive global and local information. We conducted extensive experiments in two different scenarios on two real traffic datasets. Results show that M$^3$FGM outperforms the baselines and variant models, achieves the best results in both datasets and scenarios.

\end{abstract}
\section{Introduction}
Utilizing graph structure to model spatial-temporal data in the prediction task has been popular in recent years \cite{Graphwavenet,ASTGCN,SysGCN,STODE,DD,DSTAGNN}. It is critical for various applications including traffic flow prediction, forecasting, and user activity detection.  Most of these works train models under the assumption that a massive amount of real-world spatial-temporal data can be centralized. However, with increasing concerns about data privacy and access restrictions due to existing licensing agreements and commercial competition, there are numerous real-world cases in which spatial-temporal data is decentralized. For instance, in traffic flow prediction, different organizations or companies collect traffic data by their private deployed road sensors and these data cannot exchange it due to privacy preservation or commercial reasons.

As an effective solution to data privacy protection, Federated Learning (FL) \cite{FL} has attracted significant research efforts recently. FL is a learning paradigm for model training that collaborates with clients (i.e., local data owners) without exposing their original data. By integrating all client model weights or gradients, the FL-trained model demonstrates superior generalization capabilities.

Recent research has introduced a series of FL-based models for spatial-temporal data prediction while preserving privacy \cite{PPTFP,CEFL}. However, these models do not consider the inherent spatial dependencies of the data. Current works focus on integrating FL with graph neural networks(GNNs), which can be divided into two categories: \textit{1) Client-side GNN model training for local model updates}: A common characteristic of these approaches \cite{FedCG,HAFL-GHN} is their emphasis on training with well-established graph-structured data. In practice, not all clients possess built-in graph structure datasets, which raises the question of how to process node-level data using GNNs in such contexts. \textit{2) Server-based GNN model training for enhanced FL aggregation}: Techniques such as PFL \cite{PFL} employ GCN to perform model parameters aggregation according to the clients' relational graph structure, introducing a supervised loss function with graph smoothness regularization for training both local and server models. BiG-Fed \cite{BiG-Fed} devises bi-level optimization schemes for training local models and GNN models with dual objective functions and proposes an unsupervised contrastive learning loss function. Despite these methods consider the structural relationships among clients and offer GNN-based model parameter aggregation techniques, they do not fully exploit the true capabilities of GNNs as they are unable to directly model the dependency relationships within spatial-temporal data. Consequently, their performance is significantly distant from that of centralized GNN approaches. 

In recent years, there has been an architectural approach called Split Federated Learning(SFL) that divides a complete model into several parts, placing them on the client and server sides respectively, such as \cite{sfl} and \cite{etesfl}.  This approach is primarily adopted due to the limited computational resources of the devices participating in federated learning.  However, recently, meng et.al have successfully employed this framework to enable GNNs to directly participate in spatial-temporal data processing, proposed CNFGNN \cite{CNFGNN}. Specifically, CNFGNN partitions the complete model into two components: employing identical encoder-decoder models on all clients, with the encoder used to extract local temporal embeddings, and the decoder utilized to generate predictions. Graph Network (GN) \cite{GN} is employed on the server side to obtain spatial embeddings by aggregating the local temporal embeddings uploaded from the clients. CNFGNN can be regarded as a GNN-oriented SFL method.

Nonetheless, two significant issues remain. (1) For CNFGNN, when employing trained model for inference, some clients might be unable to connect to the server due to network disconnection. While it is feasible to replace missing embeddings with all-zero data, this approach significantly diminishes predictive performance. Moreover, these offline clients cannot generate predictions without the server model. (2) The performance of GNN training relies heavily on the accuracy of the graph structure. However, the graph structure of clients in existing methods \cite{STGCN,FastGNN,CNFGNN} is constructed manually in a heuristic way, which might not represent client relations properly, leading to deteriorated performance.

In this paper, we propose a new GNN-oriented split federated learning method for spatial-temporal data prediction, named node {\bfseries M}asking and {\bfseries M}ulti-granularity {\bfseries M}essage passing-based Federated Graph Model (M$^3$FGM) to overcome the above issues. To address the concern of offline clients, we propose a MaskNode layer to the server model to simulate that clients are offline during the training phase. Additionally, we devise a dual-sub-decoders structure for the client model's decoder, permitting offline clients to make predictions during the inference phase. For the issue of graph structures, a new GNN layer, named Multi-Granularity Message Passing (MGMP) layer, is proposed. We construct a comprehensive coarse-grained graph, referred to as the cluster graph, by applying spectral clustering on the client graph. The MGMP layer empowers each client node to aggregate fine-grained local information from neighbors in the client graph and global coarse-grained information from the cluster graph.

The contribution of this paper is summarized as follows:

(1) As far as we know, this paper is the first to consider the non-ideal scenario when designing a GNN-oriented SFL method. We propose MaskNode to enhance the model robustness and design a dual-sub-decoders structure, enabling offline clients to make independent predictions.

(2) We propose a novel GNN Layer, MGMP Layer, which enables client nodes to perceive local and global information through multi-granularity message passing.

(3) We propose  M$^3$FGM for spatial-temporal data prediction under privacy protection. The extensive experiments demonstrate the effectiveness of our model on two real-world traffic datasets.

\section{Related Work}
Our method combines elements from graph neural networks, split federated learning. We now review related works in these areas and discuss their relevance to our work.

\subsection{Graph Neural Networks}
Graph Neural Networks (GNNs) have demonstrated outstanding efficacy across a diverse range of learning tasks involving graph-structured data, such as node classification \cite{semi,Nodeaug,GraphMixup}, link prediction \cite{GSNOP,chang2020continuous}, spatial-temporal data modeling \cite{peng2021dynamic,shao2022long}. Although GNNs exploit a powerful inductive bias to extract meaningful information from graph-structured data, there are challenges that need to be addressed to fully exploit their potential. One critical aspect of GNN performance is the accurate representation of graph structure. In real-world scenarios, graph structures can be highly complex, making it difficult to manually construct them using only prior knowledge. Another challenge faced by GNNs is the difficulty in capturing long-range dependencies within a limited number of message passing steps. This limitation can hinder the learning capabilities of GNNs, especially in scenarios where long-range interactions play a significant role. Therefore, in this paper, we propose a novel GNN Layer to address above issues through multi-granularity message passing.

Furthermore, most studies necessitate centralized data during training and inference processes. This reliance on centralized data leads to privacy concerns, especially when dealing with sensitive information in domains like healthcare, finance, or social networks. Consequently, there is a burgeoning interest in developing privacy-preserving GNNs that facilitate distributed learning across multiple entities, ensuring data confidentiality and compliance with data protection regulations.

\subsection{Split Federated Learning}
Federated learning (FL) \cite{FL} is a machine learning paradigm that enables multiple entities, such as mobile devices, edge nodes, or data centers, to collaboratively train a model while maintaining the privacy and decentralization of their local data. In a typical federated learning setting, multiple clients and a central server participate in training a global model. The global model is copied in multiple copies and deployed on each client. Each participating client trains the model locally using its data and sends only the updated model parameters to the central server for aggregation. 




Split learning (SL) \cite{SL} is a technique that divides a complete model into several components to enable efficient utilization of computational resources across a network of devices, leveraging their individual strengths while minimizing the overall computational burden. SL can also achieve increased scalability in large-scale distributed systems.

Recent research has integrated SL with FL to address high training latency for clients with limited resources\cite{sfl}\cite{etesfl}. This combination, referred to as Split Federated Learning (SFL), typically divides the global model into two components: client-side and server-side components. Clients access only the client-side component, while the server exclusively accesses the server-side component. In SFL, clients send their processed data (outputs from the client-side model) to the server, where the server-side model continues training. After calculating the loss and updating the gradient, the server adjusts the server-side model, and the gradients of the processed data are sent back to the clients. Clients then update the client-side model based on the gradient. By training part of the model on the server, SFL significantly reduces the computational burden for resource-constrained devices. Collaborative training between clients and the server ensures that the original data remains stored locally on the client, preventing sensitive information disclosure. However, most studies primarily focus on standard deep learning models such as CNN and RNN, with GNN-oriented SFL being rarely studied. CNFGNN \cite{CNFGNN} can be regarded as an example of a GNN-oriented SFL method. The GNN-oriented SFL method can truly unleash the potential of graph models in modeling graph-structured data, as the input for the GNN portion of the model is processed data rather than model parameters.
\label{sec:format}

\section{Problem Formulation}
\label{sec:format}

We introduce notions and definitions in this section, followed by a brief introduction to the GNN-oriented split federated learning. Let us denote the client graph constructed by the server as $G\small{=}\{V, E\}$, where $V$ is the set of client nodes, and $E$ represents the edge set. $v_i\in V$ denotes the $i$-th client node in the $G$. $N\small{=}|V|$ is the number of client nodes (the number of clients).  $c_i$ represents the client corresponding to the client node $v_i$ . Let $x_i^{t_1:t_2}$ denotes the local graph signals recorded between the timestamp $t_1$ and $t_2$ at client $i$ . $X^{t_1:t_2}$ denotes the graph signals observed at all clients between the timestamp $t_1$ and $t_2$.

The GNN-oriented split federated learning method aims to learn a client model $f_i$ for each client $c_i$, and a GNN model$f_{ser}$ for the server. At each time step $t$, each client model $c_i$  uses an encoder $f_i^{enc}$  to extract local temporal embedding $h_i^{t}$ according to $x_i^{(t-S:t)}$. Server model $f_{ser}$ computes the spatial embeddings $\{s_i^t\}_{i=1}^N$ according to $G$ and the local temporal embeddings $\{h_i^t\}_{i\small{=}1}^N$ collected from all clients. Each client model $c_i$ then uses a decoder $f_i^{dec}$ to output prediction according to $h_i^{t}$ and $s_i^{t}$. Thus, the mapping from $S$ historical graph signals $X^{(t-S):t}$ to future $T$ graph signals $X^{(t+1):(t+T)}$ can be achieved.

\begin{equation}
   [X^{(t-S):t},G]\xrightarrow{\{f_i=\{f_i^{enc},f_i^{dec}\}\}_{i=1}^N,f_{ser}}X^{(t+1):(t+T)}
 \label{eq:important}
\end{equation}

\begin{center}
\begin{figure*}[h]
\centering

  \includegraphics[width=5.0in]{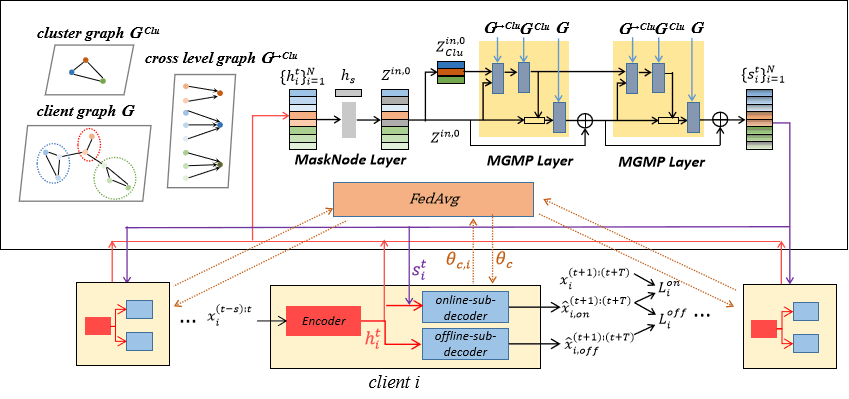}
    \caption{The overall architecture of M$^3$FGM}\label{sim_opt_eff}
\end{figure*}
\end{center}
\section{Methodology}
\label{sec:pagestyle}

Figure 1 shows the overall architecture of M$^3$FGM, and we will cover the details of the model in terms of the server model, the client model, and the training and inference process, respectively.
\subsection{Server model}

\textbf{The MaskNode Layer:} The MaskNode (MN) layer is employed exclusively during the training phase. Prior to model training, we select a mask rate $mr$. Upon feeding data into the MN layer, a certain number of client nodes, $mr \times N$, are randomly sampled. When $mr \times N$ is a noninteger, we round it down. The uploaded local temporal embeddings of these sampled client nodes are replaced with a shared trainable tensor $h_s$. The set of sampled nodes is denoted as $V_{off}$, while the set of remaining nodes is denoted as $V_{on}$. The operation of the MaskNode layer can be expressed by Eq. (2).

\begin{equation}
    \lvert \{
    \begin{array}{l}
        z_i^{in,0}=h_i^t, if \ v_i\in V_{on}\\
        z_i^{in,0}=h_{s}, if \ v_i\in V_{off}
    \end{array}
    \rvert
    \label{eq:important}
\end{equation}

$Z^{in,0}\small{=}\{z_i^{in,0}\}_{i=1}^N$ is the output of the MN layer and will be fed into the first MGMP layer. When the model training is completed and deployed, if client $c_i$ is offline, the server model will utilize the trained tensor $h_s$ as $h_i^t$ to conduct inference. Next, we briefly describe the differences between the MaskNode operation and two related techniques. Unlike the DropEdge operation \cite{DropEdge}, the MaskNode operation does not perturb the graph structure. In contrast to the masked self-supervised task \cite{GMAE}, we only replace the masked node embeddings with shared trainable tensors. We do not attempt to reconstruct or forecast the node embeddings.
\begin{center}
\begin{figure}[h]
\centering
  \includegraphics[width=2.5in]{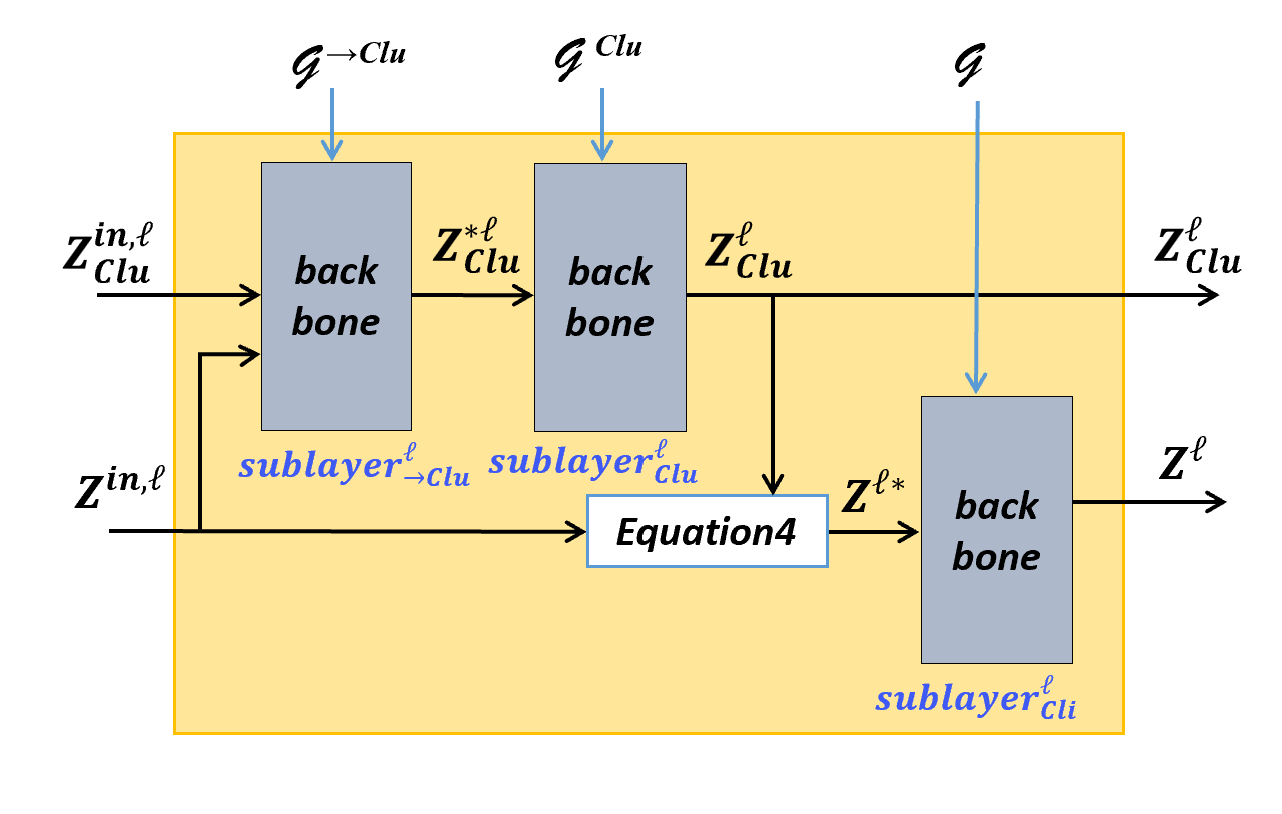}
    \caption{The structure of MGMP Layer}\label{sim_opt_eff}
\end{figure}
\end{center}

{\bfseries The MGMP layer:} The MGMP layer employs the following three graph structures for message passing: 1) client graph $G\small{=}\{V, E\}$, 2) cluster graph $G^{Clu}\small{=}\{V^{Clu}, E^{Clu}\}$, and 3) cross-level graph $G^{\rightarrow Clu}\small{=}\{V^{\rightarrow Clu},E^{\rightarrow Clu}\}$. 

The client graph $G$ is constructed manually in a heuristic way. To obtain the cluster graph $G^{Clu}$, we apply spectral clustering on the Laplacian matrix of the client graph $G$ and get $M$ clusters. Each cluster is regarded as a coarse node of the cluster graph $G^{Clu}$. Denote the set of client nodes in the $m$-th cluster as $V_m^{Clu} \subset V$. The edge of the cluster graph $G^{Clu}$ is constructed based on the client graph $G$, for example, if $V_i\in V_m^{Clu}$ and $V_j\in V_n^{Clu}$, and the $V_i$ connects to $V_j$ in client graph $G$, then the $V_m^{Clu}$ connects to $V_n^{Clu}$ in cluster graph $G^{Clu}$ . 

$G^{\rightarrow Clu}$ is a bipartite graph for accelerating the message transfers. $V^{\rightarrow Clu}\small{=}V\cup V^{Clu}$, $E^{\rightarrow Clu}$ is the edge set contains directed edges which are from client nodes in $V$ to the cluster nodes in $V^{Clu}$ corresponding to the cluster of starting node. The diagram of these three graphs is given on the left side of Fig.1.

Figure.2 shows the internal structure of the $l$-th MGMP layer, which contains three sub-layers with same backbone. The backbone can be any graph model. Let us assume that the inputs of the $l$-th layer, which include the input embeddings $Z^{in,l}\small{=}\{z_i^{in,l}\}_{i=1}^N $  of client nodes and the input embeddings $Z_{Clu}^{in,l}\small{=}\{z_{m,c}^{in,l}\}_{m=1}^M$ of cluster nodes are known. In particular, $Z_{Clu}^{in,0}$ are calculated by Eq. (3):
\begin{equation}
 z_{m,c}^{in,0}=\textstyle \sum_{v_i\in V_m^{Clu}}z_i^{in,0}/|V_m^{Clu}|,\ m=1,...,M
  \label{eq:important}
\end{equation}

We will describe how client nodes can perceive the local and global information with the help of MGMP. First, information is propagated on $G^{\rightarrow ^{Clu}}$ ($sublayer_{\rightarrow Clu}^l$) so that the cluster nodes can perceive cluster-level information from the client graph. Then, information is propagated on $G^{Clu}$($sublayer_{Clu}^l$) to obtain $Z_{Clu}^{l}\small{=} \{z_{m,c}^{l}\}_{m=1}^M$ which represents the embeddings of cluster nodes. Afterwards, the information on the cluster graph is passed back to the client graph according to Eq.(4), where $Z^{l*}\small{=}\{z_{i}^{l*}\}_{i=1}^N$ represents the embeddings of the client nodes after perceiving the coarse-grained global information, $W^l$ is a trainable matrix. Finally, fine-grained local information is propagated on $G{(sublayer_{Cli}^l)}$ to obtain the $Z^{l}\small{=}\{z_{i}^l\}_{i=1}^N$, which represents the embeddings of client nodes.

\begin{equation}
 z_i^{l*}=z_i^{in,l}||W^lz_{m,c}^l,\ v_i\in V_m^{Clu}
  \label{eq:important}
\end{equation}

{\bfseries Computational flow of server model:} The computational flow first passes the MaskNode layer and subsequently passes two MGMP layers with residual connection\cite{resnet}. Note that we only add the residual connection to the input $Z^l$ . The outputs of the last MGMP layer are sent back to clients.

\subsection{Client model}
We employ the encoder-decoder architecture on each client for the modeling of local temporal embeddings. Given an input sequence $x_i^{(t-S):t}\in R^{S\times D}$ on the $i$-th client, the encoder sequentially reads the whole sequence and outputs the hidden state $h_i^t$ as the temporal embedding of the input sequence. 

\begin{equation}
 h_i^t=f_i^{enc}(x_i^{(t-S):t})
  \label{eq:important}
\end{equation}

{\bfseries The dual-sub-decoders structure} Unlike the usual Encoder-Decoder architecture, to enable offline clients to make independent predictions, we propose the dual-sub-decoders structure. As shown in Figure 1, the dual-sub-decoders structure includes an online-sub-decoder and an offline-sub-decoder. The online-sub-decoder $f_{i,on}^{dec}$ employs local temporal embedding $h_i^t$ and spatial embedding $s_i^t$ to generate the predictions $\hat x_{i,on}^{(t+1):(t+T)}$. The offline-sub-decoder $f_{i,off}^{dec}$ only employs local temporal embedding  to output predictions $\hat x_{i,off}^{((t+1):(t+T)}$.
\begin{equation}
 \hat x_{i,on}^{(t+1):(t+T)}=f_{i,on}^{dec}(h_i^t,s_i^t)
  \label{eq:important}
\end{equation}

\begin{equation}
 \hat x_{i,off}^{((t+1):(t+T)}=f_{i,off}^{dec}(h_i^t)
  \label{eq:important}
\end{equation}

The backbone of Encoder and dual-subdecoders can be any model. In experiments, for fair, we use GRU as the backbone.

{\bfseries Loss function} To train the two sub-decoders alternately, we designed two loss functions: $L_i^on$ and $L_i^{off}$, By taking a single training sample ($x_i^{(t-S):t}$,$x_i^{(t+1):(t+T)}$) owned by client $c_i$ as an example, the two loss functions are as follows:
\begin{equation}
 L_i^{on}=\textstyle \sum_{k=1}^T(x_i^{t+k}-\hat x_{i,off}^{t+k})^2/T
  \label{eq:important}
\end{equation}
\begin{equation}
 L_i^{off}=\textstyle \sum_{k=1}^T(\hat x_{i,off}^{t+k}-\hat x_{i,off}^{t+k})^2/T
  \label{eq:important}
\end{equation}
It can be observed that $L_i^{off}$ is MSE function between the outputs of the two sub-decoders. This design aims to bring the prediction of offline-sub-decoder as close as feasible to online-sub-decoder.

\begin{algorithm}[h]
\caption{Training pipeline for $M^3FGM$ with one training sample}
\label{alg1}
\begin{multicols}{2}
\begin{algorithmic}[1]
    \REQUIRE Client graph $G$ and data ($X^{(t-S):t},X^{(t+1):(t+T)}$). Initial each client model weights as $\theta_c=\{\theta^{enc},\theta^{dec}_{on}, \theta^{dec}_{off}\}$, initial server model weights $\theta_{server}$. Initial spatial embeddings $\{s_i^t\}_{i=1}^N=s_0$, $s_0$ is a zero-valued vector. Masknode rate $mr$.
    \ENSURE Trained client model weights $\theta_c$, trained server model weights $\theta_{server}$
    \FOR {global training round $r_g=1,2,...R_g$}
    \STATE\emph{\textcolor{red}{\textbf{Step 1:}}}
        \FOR{client $i (i=1,...,N)$ in parallel}
            \FOR{local training round $r_c=1,2,...R_c$}
                \STATE $f_i^{enc}(x_i^{(t-S):t})\rightarrow h_i^t$
                \STATE $f_{i,on}^{dec}(h_i^t,s_i^t)\rightarrow \hat{x}_{i,on}^{(t+1):(t+T)}$
                \STATE $f_{i,off}^{dec}(h_i^t)\rightarrow \hat{x}_{i,off}^{(t+1):(t+T)}$
                \STATE Calculate $L_i^{on}$ and $L_i^{off}$ according to Equations (8) and (9)
                \STATE update $\{\theta^{enc}_i,\theta^{dec}_{i,on}\}$ according to $L_i^{on}$ 
                \STATE update $\theta^{dec}_{i,off}$ according to $L_i^{off}$
            \ENDFOR
        \ENDFOR
        \STATE\emph{\textcolor{red}{\textbf{Step 2:}}}
        \STATE Send latest embedding $\{h_i^t\}_{i=1}^N$ and the client model weights $\{\theta_{c,i}\}_{i=1}^N$ to the server
        \STATE Fix all client models' weights.
        \FOR{server training round $r_s=1,2,...R_s$}
            \STATE construct cluster graph $G^{clu}$ and cross-level graph $G^{\rightarrow Clu}$ according to client graph $G$
            \STATE $f_{server}(\{h_i^t\}_{i=1}^N, G, G^{clu}, G^{\rightarrow Clu}, mr)\rightarrow \{s_i^t\}_{i=1}^N$
            \STATE send $\{s_i^t\}_{i=1}^N$ to corresponding clients
            \STATE $f_{i,on}^{dec}(h_i^t,s_i^t)\rightarrow \hat{x}_{i,on}^{(t+1):(t+T)}, i=1,...,N$
            \STATE Calculate $\sum_{i=1}^N L_i^{on}$ and update $\theta_{server}$
        \ENDFOR
    \STATE\emph{\textcolor{red}{\textbf{Step 3:}}}
        \STATE Update latest graph embedding $\{s_i^t\}_{i=1}^N$
        \STATE Use FedAvg to aggregate $\{\theta_{c,i}\}_{i=1}^N\rightarrow \theta_c$
        \STATE $\{s_i^t\}_{i=1}^N$ is send to corresponding clients respectively and $\theta_c$ is send to all clients as the new model weights for next global training round.
    \ENDFOR
\end{algorithmic}
\end{multicols}
\end{algorithm}

\subsection{Training and inference process}
{\bfseries Training step} We use the alternating training method proposed in \cite{CNFGNN} to train our model to reduce communication consumption. The training and inference process of M$^3$FGM is slightly different from \cite{CNFGNN} because of the dual-subdecoders architecture. Here we briefly describe the training process:

\textit{Step 1:} Initially, the clients' models are trained for $R_c$ round with the server model and spatial embeddings fixed. Taking client $i$ for example, in each round, the offline-sub-decoder$f_{i,off}^{dec}$ is fixed, and the encoder $f_{i}^{enc}$ and online-sub-decoder $f_{i,on}^{dec}$ are trained by minimizing $L_i^{on}$ . Then the $f_{i}^{enc}$ and $f_{i,off}^{dec}$ are fixed and the $f_{i,off}^{dec}$ is trained by minimizing $L_i^{off}$. 

\textit{Step 2:} After completing $R_c$ round, all clients' model parameters $\{\theta _{c,i}\}_{i=1}^N$  and local temporal embeddings $\{h_i^t\}_{i=1}^N$ are uploaded to the server, and then the training of the server model begins for $R_s$ rounds with clients' model fixed. $\sum_{i=1}^N L_i^{on}$ is used to update the server model. 

\textit{Step 3:} Once the server model is trained, the FedAvg algorithm \cite{FL} is employed by the server to aggregate $\{\theta _{c,i}\}_{i=1}^N$ to obtain $\theta _c$. The server subsequently sends $\theta _c$ back to all clients and spatial embeddings $\{s_i^t\}_{i=1}^N$ are returned to their corresponding clients.

The above process is repeated $R_g$ times.

{\bfseries Inference step} If a client can connect to the server, the client feeds its local data to the encoder to obtain local temporal embedding, then upload embedding to the server to compute spatial embedding. After that, the client receives the spatial embedding transmitted back by the server and makes predictions using the online-sub-decoder. Conversely, when a client is unable to establish a connection to the server, it utilizes the encoder and offline-sub-decoder independently to make predictions.

\section{EXPERIMENTS}




\subsection{Datasets} Traffic data are commonly in the format of spatial-temporal graphs. We verify M$^3$FGM on two real-world traffic datasets: METR-LA and PEMS-BAY, which are released by Li et al. \cite{DCRNN}. (1)METR-LA: which records traffic speed information collected from 207 loop detectors in the highway of Los Angeles County over 4 months. (2)PEMS-BAY: which contains 6 months of traffic speed information ranging on 325 sensors in the Bay Area. 

For both two datasets, the readings of sensors are aggregated into 5 minutes windows. We standardize the data by removing the mean and scaling to unit variance. And then we split 70\% into training set, 20\% into testing set and 10\% into validation set, in chronological order. And We adopt the same data pre-processing method as \cite{CNFGNN}.

\subsection{Compared models and Settings}
Since our primary focus is on the architecture of the federated graph, rather than the specific models, we have not made comparisons with some SOTA centralized spatial-temporal graph methods. We follow the setup of \cite{CNFGNN}, comparing M$^3$FGM with four baselines. We compare M$^3$FGM with 4 baselines: GRU, GRU+ FedAvg, GRU+FMTL \cite{FMTL} and CNFGNN. These baselines all use the GRU-based encoder-decoder model \cite{GRU} as the client-side model. For each baseline, there are 2 variants of the GRU model to show the effect of on-device model complexity: one with 63K parameters and the other with 727K parameters. For CNFGNN, the encoder-decoder model on each client has 64K parameters and the GN model has 1M parameters. The experimental results of the baseline models, as reported in \cite{CNFGNN}, are utilized in the subsequent analysis. Additionally, to facilitate an objective ablative analysis, two variant models have been constructed: \textit{CNFGNN+MN}: Add the MaskNode layer to the server model of CNFGNN.  \textit{M$^3$FGM w/o MN}: M$^3$FGM without MaskNode layer.

We conduct experiments under two scenarios to verify the effectiveness of our model: an ideal scenario in which all nodes are online during the inference phase and a non-ideal scenario in which some nodes are offline during the inference phase. To ensure fair evaluation and comparison, GRU is used as the backbone of the encoder and sub-decoder in the client model when implementing M$^3$FGM and M$^3$FGM w/o MN. To optimize the model, the Adam optimizer is employed with a learning rate set at 1e-3. The root mean squared error (RMSE) metric is utilized to evaluate the predictive performance.

\begin{table*}[h]
\centering
\caption{Comparison of performance with the Rooted Mean Squared Error (RMSE) as the evaluation metrics.}\label{tab:aStrangeTable}
\renewcommand\arraystretch{1.2}
\setlength{\tabcolsep}{3.3mm}{
\begin{tabular}{ccc}
\hline
\multicolumn{1}{c}{Method}      & PEMS-BAY & \multicolumn{1}{c}{METR-LA} \\ \hline
GRU(central,63k)     & 4.124    & 11.730      \\
GRU(central,727k)     & 4.128    & 11.787      \\
GRU+GN(central,64k+1M)             & 3.816    & 11.471                       \\
                   \hline
GRU(local,63k)                    & 4.010    & 11.801   \\
GRU(local,727k)                    & 4.152    & 12.224   \\
              
GRU(63k)+FedAvg             & 4.512    & 12.132     \\
GRU(727k)+FedAvg          & 4.432    & 12.058     \\

GRU(63k)+FedMTL             & 3.9561    & 11.548    \\
GRU(727k)+FedMTL             & 3.955    & 11.570    \\
CNFGNN(64k+1M)                  & 3.822    & 11.487  \\
CNFGNN(64k+1M) +MN                  & 3.831    & 11.504  \\

M$^3$FGM w/o MN      & 3.697    & 11.371  \\ 
M$^3$FGM ($mr$=25\%)     & \textbf{3.684}    & \textbf{11.352}  \\ \hline
\end{tabular}
\vspace{-1em}}
\end{table*}

\begin{table*}[]
\centering
\caption{Performance comparison under the non-ideal scenario}
\renewcommand\arraystretch{1.2}
\setlength{\tabcolsep}{3mm}{

\begin{tabular}{c|c|c|c}
\hline
                          & \multicolumn{1}{c|}{$\textit{online}\vert \textit{offline}$} & \multicolumn{1}{c|}{$75\%\vert 25\%$} & \multicolumn{1}{c}{$65\%\vert35\%$} \\ \hline
\multirow{4}{*}{\rotatebox{270}{PEMS-BAY}}
                        & GRU(local)                                   & 4.010                        & 4.010                       \\ 
                          & CNFGNN                                   & $3.972\vert*$                       & $4.232\vert*$                        \\ 
                         & CNFGNN+MN($mr=10\%$)                          & $3.904\vert*$                       & $4.163\vert*$                        \\ 
                          & M$^3$FGM w/o MN                             & $3.837\vert3.967$                   & $4.021\vert3.969$                    \\ 
                          & M$^3$FGM ($mr=25\%$)                             & $3.741\vert3.934$                    & $3.836\vert3.938 $                   \\ \hline
\multirow{4}{*}{\rotatebox{270}{METR-LA}}
                        & GRU(local)                                   &  11.801                        &  11.801                       \\ 
                          & CNFGNN                                   & $11.637\vert* $                      & $11.809\vert*$                       \\ 
                         & CNFGNN+MN($mr=10\%$)                          & $11.563\vert*  $                     & $11.704\vert* $                      \\ 
                          & M$^3$FGM w/o MN                             & $11.516\vert11.787$                  & $11.633\vert11.788 $                 \\ 
                          & M$^3$FGM($mr=25\%$)                             & $11.423\vert11.782$                 & $11.513\vert11.782$                  \\ \hline
\end{tabular}
}
\end{table*}

\subsection{Performance comparison under the ideal scenario}
Table 1 reveals that M$^3$FGM achieves the lowest prediction error on both datasets. Specifically, M$^3$FGM and CNFGNN demonstrate superior performance compared to GRU+FedAvg and GRU+ FedMTL by taking into account the spatial correlation of client nodes.

{\bfseries Ablation analysis:} In Table 1: 1) M$^3$FGM outperforms M$^3$FGM w/o MN, indicating that the MaskNode layer contributes to enhanced prediction performance under the ideal scenario. 2) M$^3$FGM w/o MN surpasses CNFGNN. Given that the client models of the two methods share the same structure under the ideal scenario, this result suggests that the MGMP Layer is instrumental in improving prediction performance. 3) CNFGNN+MN exhibits slightly inferior performance compared to CNFGNN on both datasets. We hypothesize that this is because the server model of CNFGNN+MN struggles to aggregate valuable information within a few message-passing steps when certain node embeddings are masked. In contrast, M$^3$FGM with the MGMP layer addresses this issue by passing neighbor and global information.

\subsection{Performance comparison under the non-ideal scenario}

To simulate non-ideal situations, we set two different client offline rates: 25\% and 35\%. We calculate the RMSE of the models separately for online and offline nodes. Given that each client of GRU (local) makes predictions independently during the inference phase, we also compare M$^3$FGM with GRU (local). We find that the prediction performance of M$^3$FGM on offline clients surpasses that of GRU (local) on all clients. The results show that the training method adopted by M$^3$FGM enables the offline-sub-decoder, which is used for local independent prediction, to outperform GRU(local). Moreover, M$^3$FGM exhibits better robustness than CNFGNN as the offline rate increases. Comparing the prediction error of M$^3$FGM with CNFGNN on online nodes, the increase rate of RMSE is 2.5\% vs. 6.5\% on PEMS-BAY and 0.8\% vs. 1.5\% on METR-LA.

{\bfseries Ablation analysis:} Upon analyzing the results in Table 2, we observe that 1) CNFGNN+MN outperforms CNFGNN on online clients, and M$^3$FGM surpasses M$^3$FGM w/o MN on both online and offline clients. These results demonstrate that the MaskNode layer improves the model's robustness. 2) Comparing the experimental results of M$^3$FGM and CNFGNN+MN on online clients, we deduce that employing the MGMP layer enhances the prediction performance on online clients under the non-ideal scenario. This finding highlights the importance of incorporating the MGMP layer in non-ideal scenarios to achieve improved prediction accuracy and model robustness.

\begin{figure*}[h]
\centering
    \subfigure[METR-LA]{
    \begin{minipage}{0.22\linewidth}
	{
		  \includegraphics[scale=0.15]{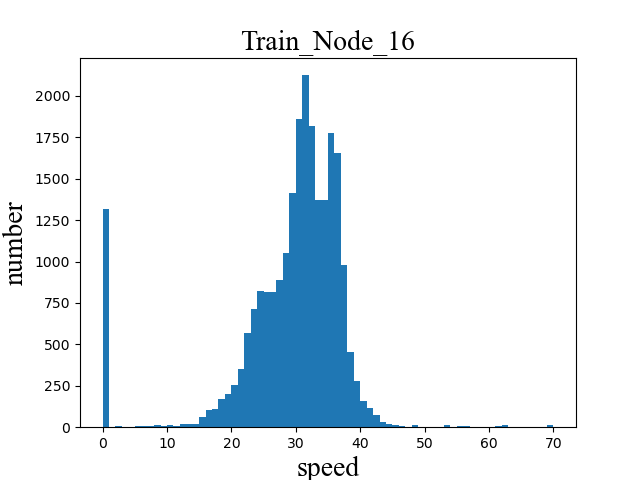}   
            \includegraphics[scale=0.15]{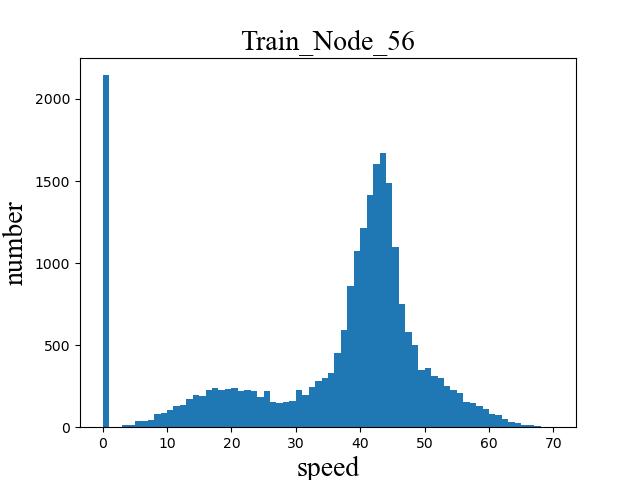}
       
            \includegraphics[scale=0.15]{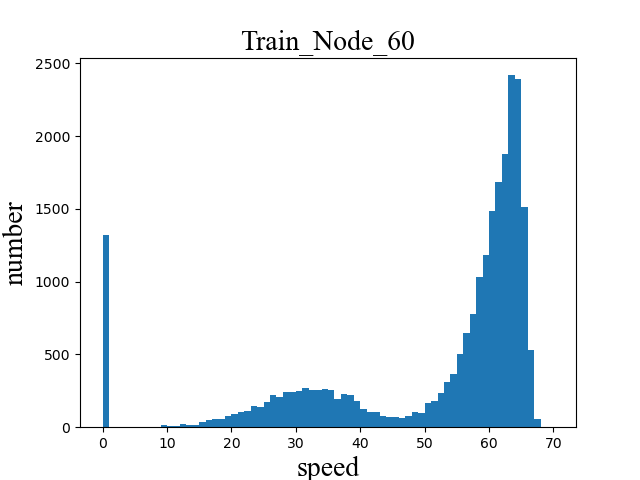}
       
            \includegraphics[scale=0.15]{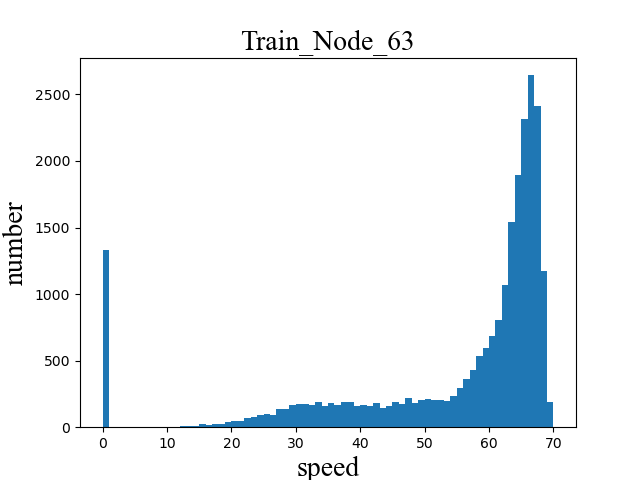}
      
            \includegraphics[scale=0.15]{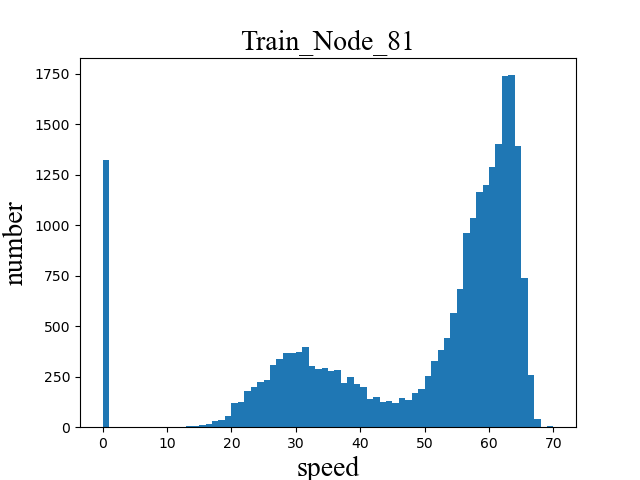}
        
            \includegraphics[scale=0.15]{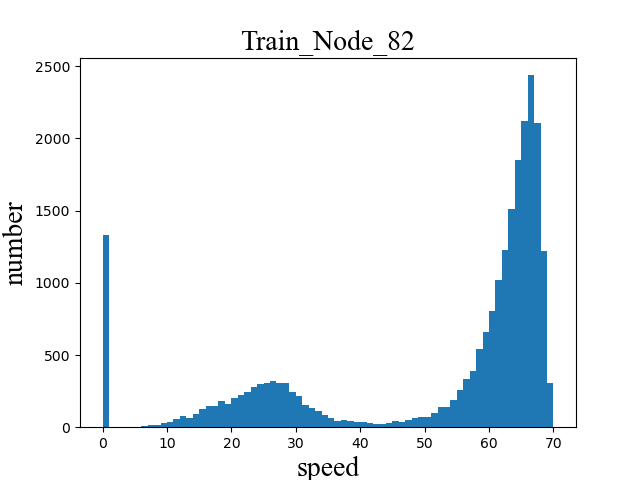}
    
	}
    \end{minipage}
    \begin{minipage}{0.22\linewidth}
	{
		  \includegraphics[scale=0.15]{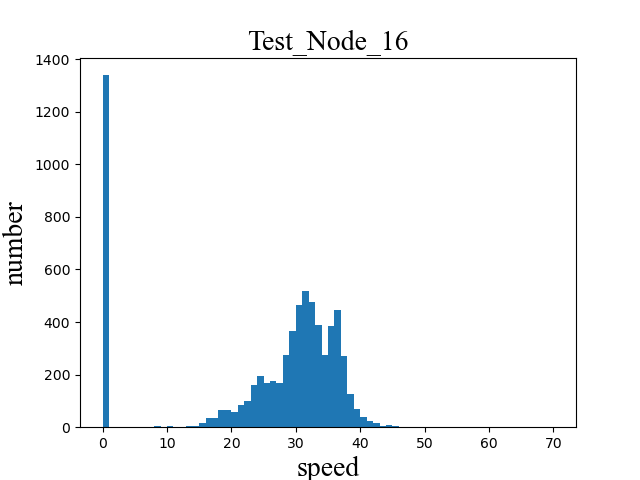}   
            \includegraphics[scale=0.15]{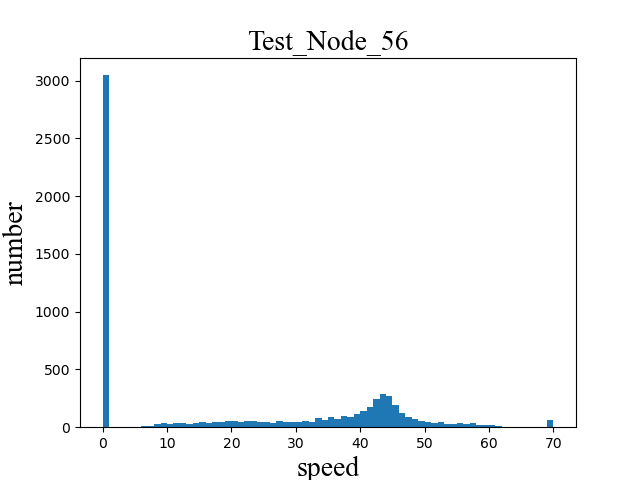}
      
            \includegraphics[scale=0.15]{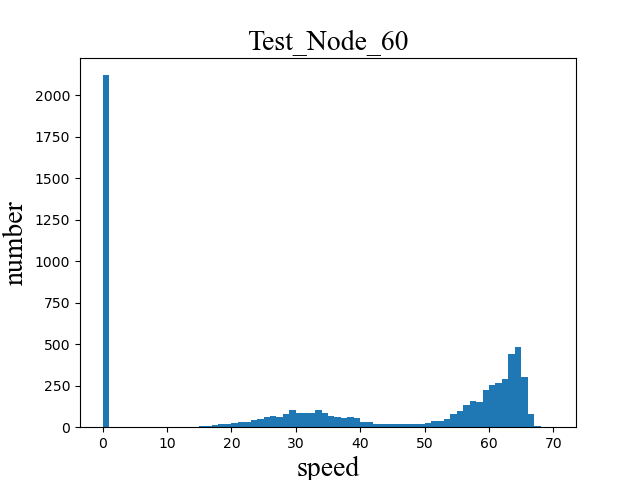}
        
            \includegraphics[scale=0.15]{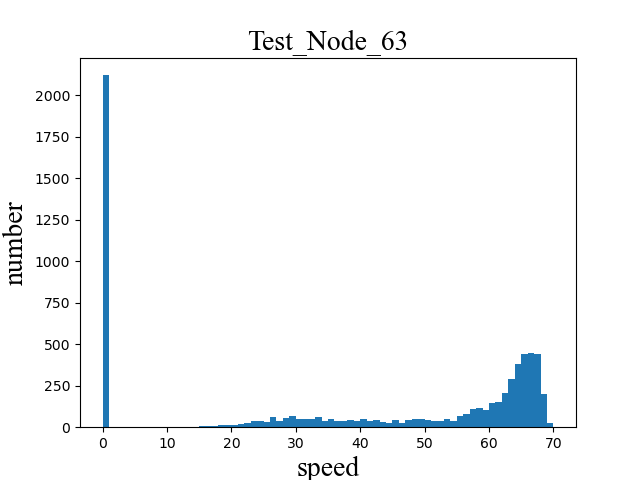}
        
            \includegraphics[scale=0.15]{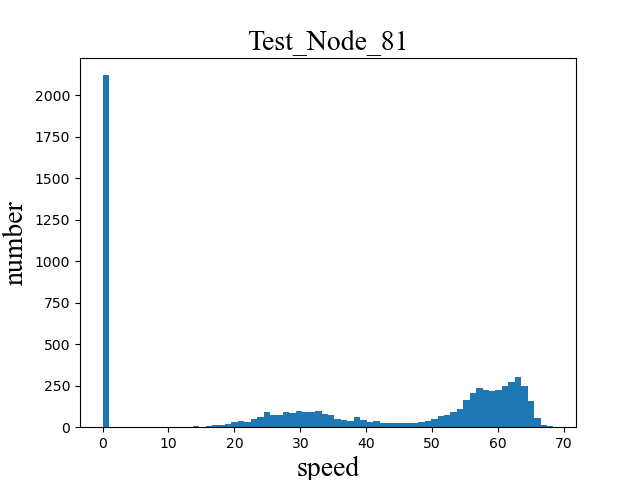}
        
            \includegraphics[scale=0.15]{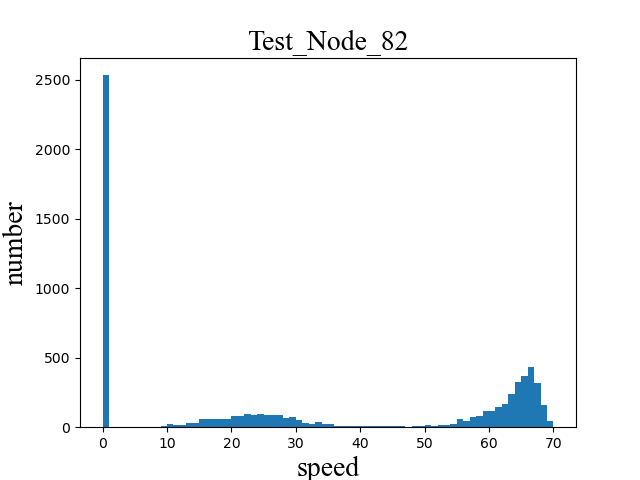}
	}
    \end{minipage}}
    \subfigure[PEMS-BAY]{
    \begin{minipage}{0.22\linewidth}
	{
		\includegraphics[scale=0.15]{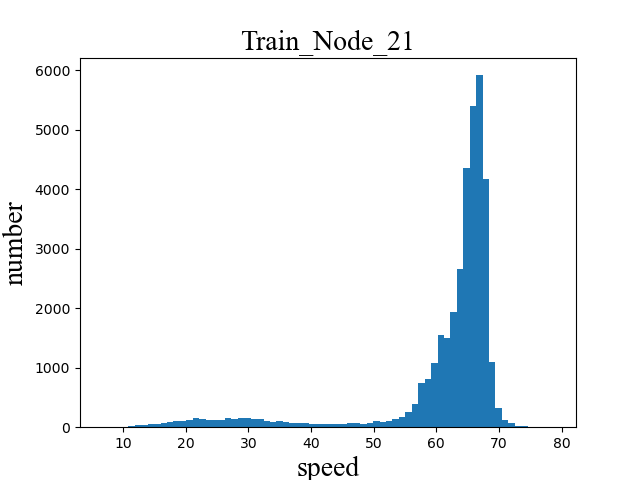}   
            \includegraphics[scale=0.15]{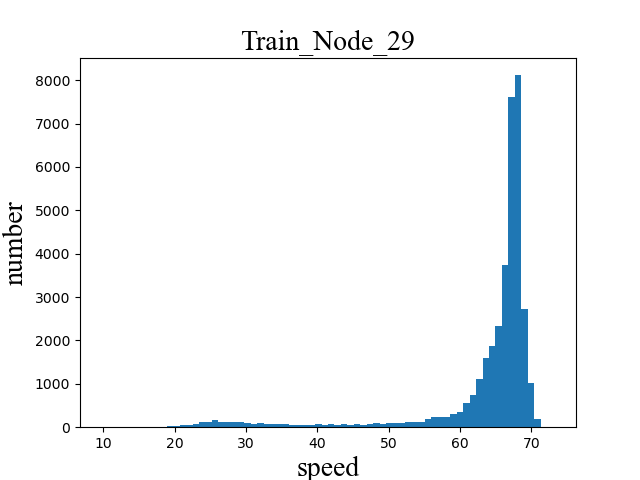}
            \includegraphics[scale=0.15]{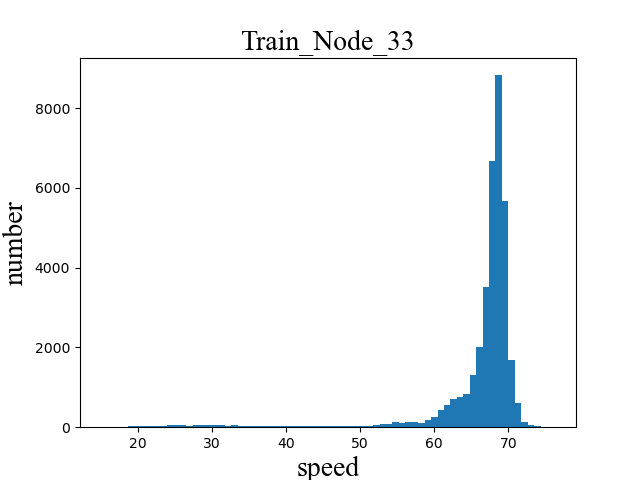}
            \includegraphics[scale=0.15]{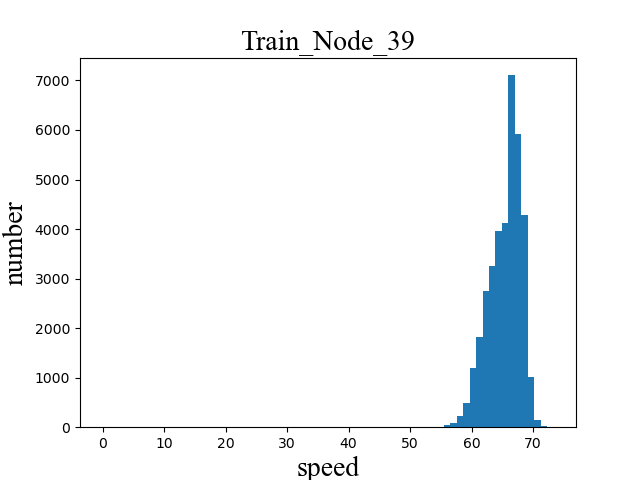}
            \includegraphics[scale=0.15]{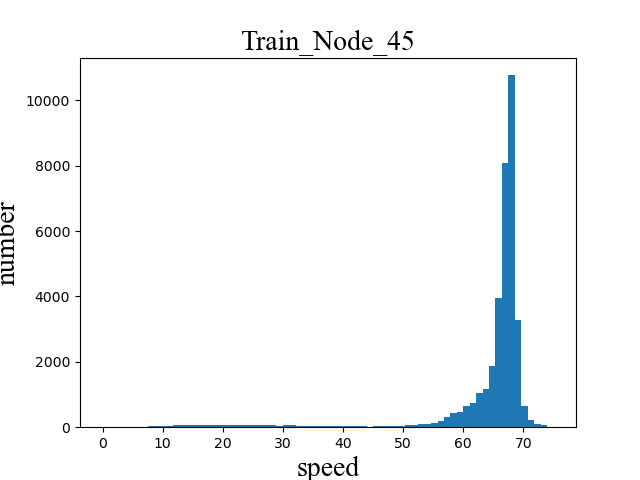}
            \includegraphics[scale=0.15]{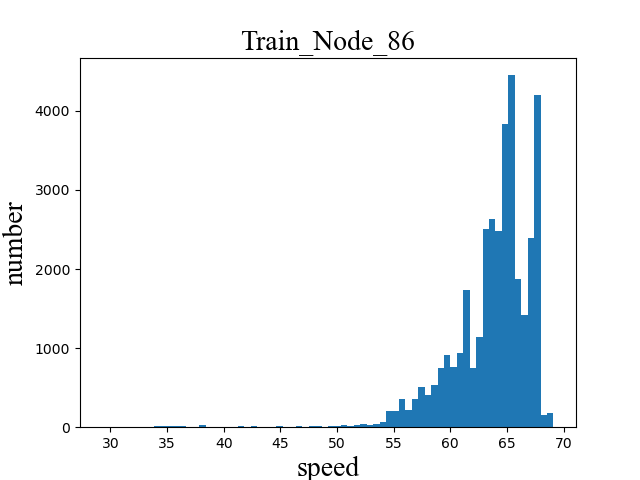}
	}
    \end{minipage}
    \begin{minipage}{0.22\linewidth}
	{
		  \includegraphics[scale=0.15]{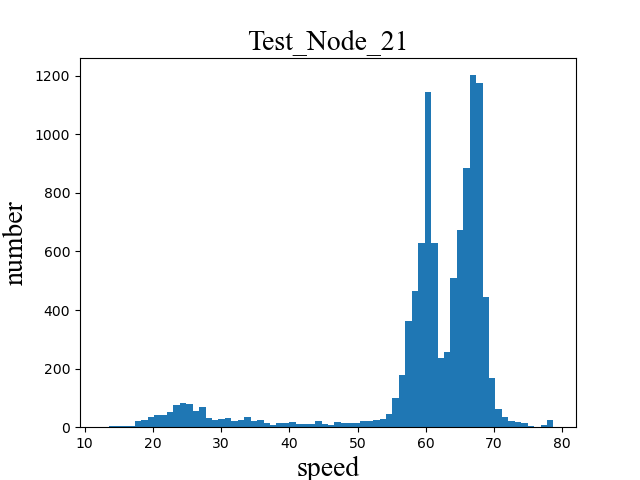}   
            \includegraphics[scale=0.15]{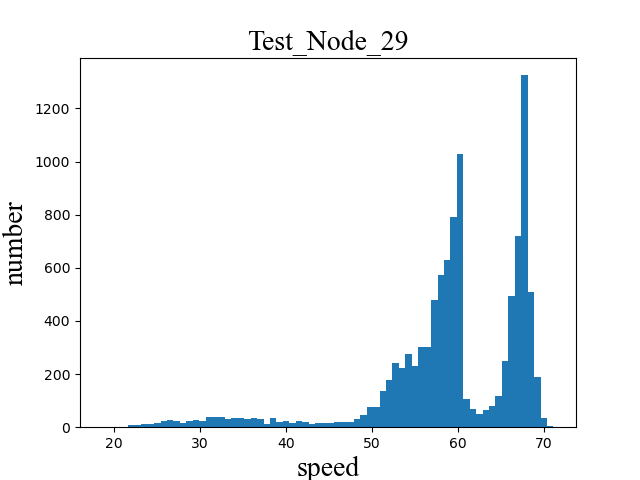}
            \includegraphics[scale=0.15]{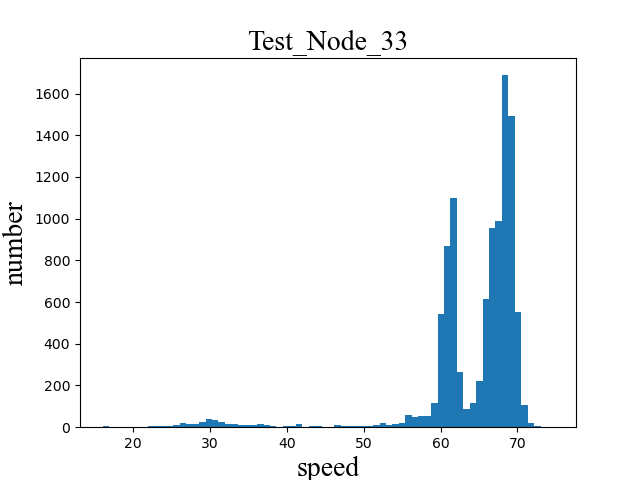}
            \includegraphics[scale=0.15]{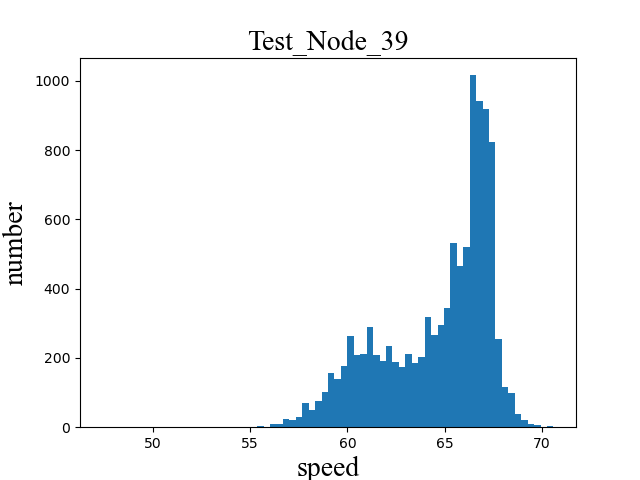}
            \includegraphics[scale=0.15]{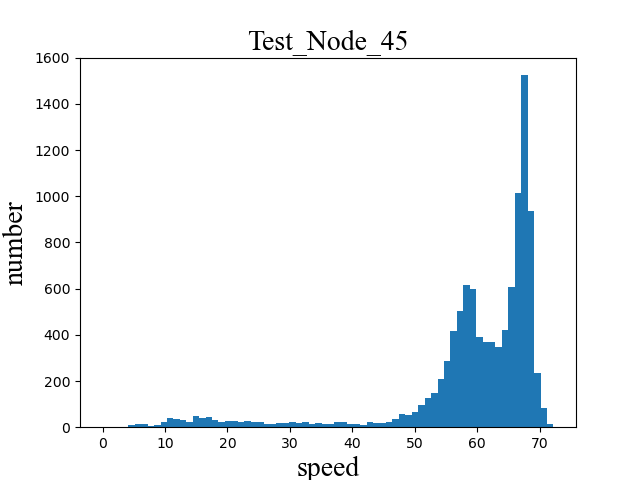}
            \includegraphics[scale=0.15]{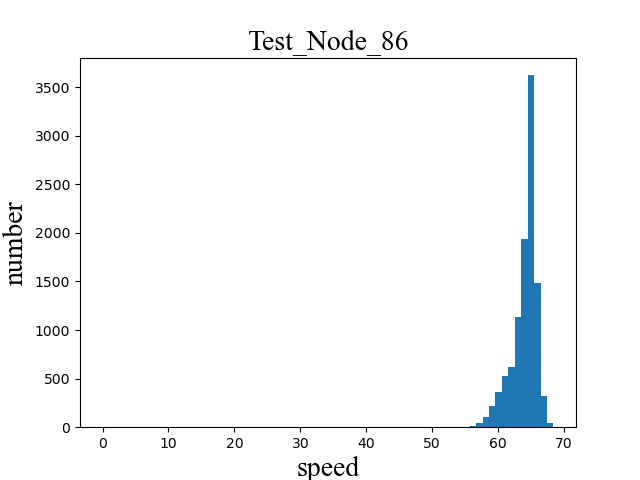}
	}
    \end{minipage}}
    
    \caption{Data distribution of training set data and test set data}
    \vspace{-2em}
\end{figure*}

\begin{figure*}[h]
\centering
    \subfigure[METR-LA] 
	{
		  \includegraphics[scale=0.25]{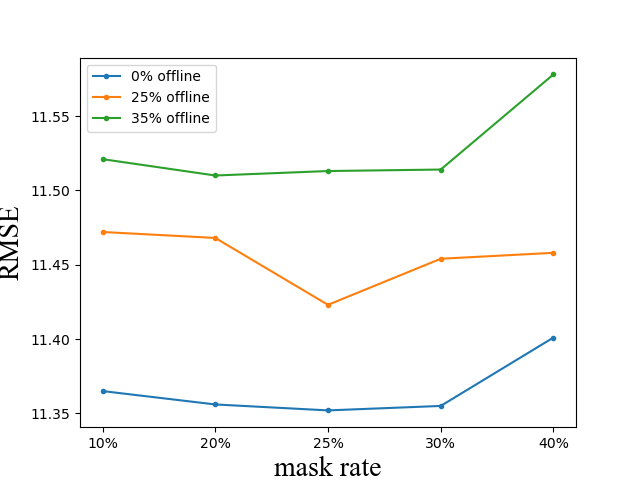}   
	}
    \subfigure[PEMS-BAY] 
	{
			
		  \includegraphics[scale=0.25]{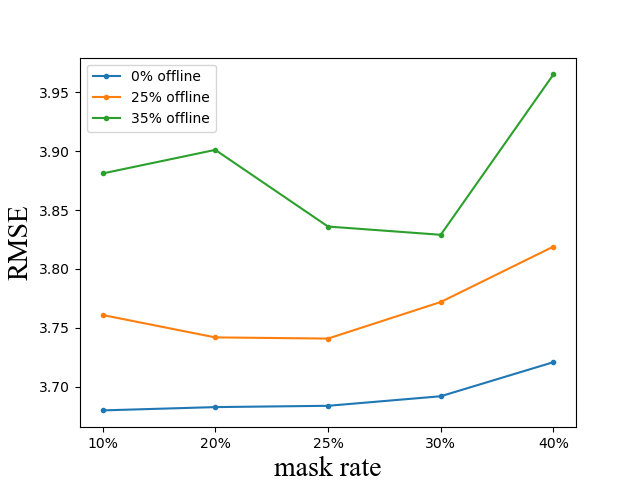}   
	}

      \caption{Comparison of performance under different mask rate and offline rate with RMSE.}\label{tab:aStrangeTable}
      \vspace{-2em}
\end{figure*}

\subsection{Effect of mask node rate and Discussion}
In order to investigate the effect of mask rate on model prediction performance, we selected five mask rates to train M$^3$FGM: 10\%, 20\%, 25\%, 30\%, and 40\%, and conducted inference on two datasets under different offline rates:0\%, 25\%, 35\%. Figure. 4 displays the performance of the model on online nodes. From these results, it can be observed that: (1) On the two datasets, it is not the case that the lower or higher the mask rate, the better. When the offline rate is fixed, compared to other mask rates, selecting a mask rate closer to the offline rate leads to better performance of the model. When the offline rate is 0\%, which is the ideal scenario, choosing a mask rate within the range of 10\% to 25\% would be better. (2) When the mask rate is fixed, as the offline rate increases, the performance of model decreases. (3) The prediction error of the model on the PEMS-BAY dataset is significantly lower than that on the METR-LA dataset. However, the model's performance on the PEMS-BA dataset exhibits greater fluctuations with mask rate variation compared to its performance on the METR-LA dataset.


To understand the underlying principles of these trends, we analyzed the data used in the experiments. We selected six nodes from the first 100 nodes ranked by their IDs in METR-LA and PEMS-BAY and illustratedthe statistical histogram of traffic speed of training data and test data of different nodes of METR-LA and PEMS-BAY in Figure. 3(a) and Figure. 3(b), respectively. The analysis revealed the following key insights: (1)On the METR-LA dataset, the histograms show that the data distribution varies with nodes, and most importantly, their training and test data distributions exhibit considerable discrepancies. (2)On the PEMS-BAY dataset, however, the differences between the training and test data distributions are much smaller. Additionally, the data distributions among different nodes are more similar to each other. 

Based on this analysis, we can conclude that on the METR-LA dataset, existing a strong shift in data distribution. The occurrence of data distribution shift can result in a significant decline in the predictive performance of a model. For instance, when employing traffic forecasting model trained on the data collected in sunny days for rainy or foggy environments, inevitable performance drop can often be observed in such scenarios. Because the trained models tend to overfit the training data and show vulnerability to the statistic changes at testing time, substantially limiting the generalization ability of the learned representations. Thus, selecting an appropriate mask rate can effectively prevent model overfitting and reduce prediction errors.  In contrast, The data distribution among various nodes in the PEMS-BAY dataset is relatively similar, and the differences between the training data distribution and the testing data distribution within each node are not substantial. This suggests that the correlation among nodes in the PEMS-BAY dataset is stronger, resulting in a more significant impact of the offline rate and mask rate on the model's performance. These observations above emphasize the importance of selecting an appropriate mask rate based on the specific characteristics of the dataset to achieve optimal model performance.

\section{CONCLUSION}
\label{sec:majhead}
In this paper, we propose a new GNN-oriented split federated learning method, named node Masking and Multi-granularity Message passing-based Federated Graph Model(M$^3$FGM) specifically developed for spatial-temporal data prediction in scenarios where data decentralization is imperative due to privacy concerns. We improve robustness of model by introducing the MaskNode layer  and the proposed dual-sub-decoders structure enables independent offline prediction. In addition, a new GNN layer named Multi-Granularity Message Passing (MGMP) layer enables each client node to perceive global and local information in a short message passing steps. We conducted evaluations under both ideal and non-ideal scenarios, the comprehensive experimental results demonstrate the superiority of the proposed M$^3$FGM model in comparison to existing methods in terms of prediction accuracy and robustness under various conditions. 

\section{Acknowledgments}
This work was supported by the National Science Foundation of China 61562041.
%
%
%
%

\bibliographystyle{splncs04} 
\bibliography{refs} 

\end{document}